\documentclass{elsarticle}

\usepackage{tabularx}
\usepackage{multirow}
\usepackage{amssymb}
\usepackage{amsmath}
 \usepackage{url}
 \usepackage{xcolor}
 \usepackage{colortbl}
 \usepackage{subcaption}
 \usepackage{subfloat}
 \usepackage{bm}
 \usepackage{enumitem}

\newlength\xlength
\newlength\mylength
\newlength\ylength
\definecolor{Gray}{gray}{0.8}

\definecolor{newcolor}{rgb}{.8,.349,.1}

\journal{Expert Systems with Applications}

\begin{document}

\title{Detecting Unseen Falls from Wearable Devices using Channel-wise Ensemble of Autoencoders}

\author[uft,uhn]{Shehroz S. Khan\corref{cor1}} 
\cortext[cor1]{Corresponding author: 
  Tel.: +1 416-597-3422;  
  Fax: +1 416-597-6201;}
\ead{shehroz.khan@utoronto.ca}
\author[uhn]{Babak Taati
\ead{babak.taati@uhn.ca}}

\address[uhn]{Toronto Rehabilitation Institute, 550 University Ave, Toronto, ON, M5G 2A2, Canada}
\address[uft]{University of Toronto, Canada}

\begin{abstract}
A fall is an abnormal activity that occurs rarely, so it is hard to collect real data for falls. It is, therefore, difficult to use supervised learning methods to automatically detect falls. Another challenge in using machine learning methods to automatically detect falls is the choice of engineered features. In this paper, we propose to use an ensemble of autoencoders to extract features from different channels of wearable sensor data trained only on normal activities. We show that the traditional approach of choosing a threshold as the maximum of the reconstruction error on the training normal data is not the right way to identify unseen falls. We propose two methods for automatic tightening of reconstruction error from only the normal activities for better identification of unseen falls. We present our results on two activity recognition datasets and show the efficacy of our proposed method against traditional autoencoder models and two standard one-class classification methods.
\end{abstract}

\maketitle

\section{Introduction}


Falls are a major cause of both fatal and non-fatal injury and a hindrance in living independently. 
Each year an estimated $424, 000$ individuals die from falls globally and $37.3$ million falls require medical attention \citep{website:whofall1}.
Experiencing a fall may lead to a fear of falling~\citep{igual2013challenges}, which in turn can result in lack of mobility, less productivity and reduced quality of life. 
There exist several commercial wearable devices to detect falls~\citep{pannurat2014automatic}; most of them use accelerometers to capture motion information. They normally come with an alarm button to manually contact a caregiver if the fall is not detected by the device. However, most of the devices for detecting falls produce many false alarms \citep{el2013fall}. Automatic detection of falls is long sought; hence, machine learning techniques are needed to automatically detect falls based on sensor data. However, a fall is a rare event that does not happen frequently \cite{Stone2015Fall,khan2014iwaal}; therefore, during the training phase, there may be very few or no fall samples. Standard supervised classification techniques may not be suitable in this type of skewed data scenario. 
Another issue regarding the use of machine learning methods in fall detection is the choice of features. Traditional activity recognition and fall detection methods extract a variety of domain specific features from raw sensor readings to build classification models~\citep{Ravi:2005,khan2016classification}. It is very difficult to ascertain the number or types of features, specially in the absence of fall specific training data to build generalizable models. 

To handle the problems of lack of training data from real falls and the difficulty in engineering appropriate features, we explore the use of Autoencoders (AE) that are trained only on normal activities. AEs can learn generic features from the raw sensor readings and can be used to identify unseen falls as abnormal activities during testing based on a threshold on the reconstruction error. 
We present two ensembles approaches of AE that train on the raw data of the normal activities from different channels of accelerometer and gyroscope separately and the results of each AE is combined to arrive upon a final decision. 
 Typically, while using AE, the maximum of reconstruction error on the training set is considered as the threshold to identify an activity as abnormal. However, we experimentally show that such threshold may not be appropriate for detecting falls due to noisy sensor data. 
 We present two threshold tightening techniques to remove few outliers from the normal data. Then, either a new threshold is derived using inter-quartile range or by training a new AE on the training data with outliers removed.
 We show result on two activity recognition datasets that contain different normal activities along with falls from wearable sensors. 

The rest of the paper is organized as follows. In the next Section, we present a brief introduction to Autoencoders. Section \ref{sec:related} reviews the literature on fall detection using AE and on the use of AE in general outlier detection tasks. We present the proposed channel-wise ensemble of autoencoder and two threshold tightening approaches using reconstruction error in Section \ref{sec:threshold}. Experimental analysis and results are discussed in Section \ref{sec:results}, followed by conclusions and future work in Section \ref{sec:conclusions}.

\section{Brief Introduction to Autoencoders}
\label{sec:ae}
An AE is an unsupervised multi-layer neural network that learns compact representation of the input data~\citep{scholz2002nonlinear}. An AE tries to learn an identity function such that its outputs are similar to its inputs. However, by putting constraint on the network, such as limiting the number of hidden neurons, it can discover compact representations of the data that can be used as features for other supervised or unsupervised learning tasks. An AE is often trained by using the backpropagation algorithm 
 and consists of an encoder and decoder part. If there is one hidden layer, an AE takes the input $\textbf{x}\in \mathbb{R}^d$ and maps it onto $\textbf{h}\in \mathbb{R}^p$, s.t.

\begin{equation}
  \textbf{h}=f(W\textbf{x}+b)
\end{equation}

\noindent where $\textbf{W}$ is a weight matrix and $\textbf{b}$ is a bias term and $f(.)$ is a mapping function.
This step is referred to as encoding or learning latent representation, after which $\textbf{h}$ is mapped back to reconstruct $\textbf{y}$ of the same shape as $\textbf{x}$, i.e.

\begin{equation}
  \textbf{y}=g(\textbf{W$'$h}+\textbf{b$'$})
\end{equation}

This step is referred to as decoding or reconstructing the input back from latent representation. An AE can be used to minimize the squared reconstruction error, $L$ i.e.,

\begin{equation}
\label{eq:re}
  L(\textbf{x,y})=\parallel \textbf{x -y} \parallel ^2
\end{equation}

AE can learn compact and useful features if $p<d$; however, it can still discover interesting structures if $p>d$. This can be achieved by imposing a sparsity constraint on the hidden units, s.t. neurons are inactive most of the time or the average activation of each hidden neuron is close to zero. To achieve sparsity, an additional sparsity parameter is added to the objective function. 
%
%
%
Multiple layers of AEs can be stacked on top of each other to learn hierarchical features from the raw data. They are called Stacked AE (SAE). During encoding of a SAE, the output of first hidden layer serves as the input to the second layer, which will learn second level hierarchical features and so on. For decoding, the output of the last hidden layer is reconstructed at the second last hidden layer, and so on until the original input is reconstructed.
%

\section{Related Work}
\label{sec:related}
AEs can be used both in supervised and unsupervised mode for identifying falls. In a supervised classification setting, AE is used to learn representative features from both the normal and fall activities. This step can be followed by a standard machine learning classifier trained on these compressed features~\citep{li2014unsupervised} or by a deep network~\citep{jokanovic2016radar}. In the unsupervised mode or One-Class Classification (OCC)~\citep{Khan:KER:2014} setting, only  data for normal activities is present during training the AE. In these situations, an AE is used to learn representative features from the raw sensor data of normal activities. This step is followed by either employing (i) a discriminative model by using one-class classifiers or (ii) a generative model with appropriate threshold based on reconstruction error, to detect falls and normal activities. The present paper follows the unsupervised AE approach with a generative model and finding an appropriate threshold to indentify unseen falls.

A lot of work has been done in evaluating the feasibility of learning generic representations through AEs for general activity recognition and fall detection tasks. Pl\"{o}tz et al.~\citep{Plotz2011Feature} explore  the potential  of  discovering universal features for context-aware application using wearable sensors. They present several feature learning approaches using PCA and AE and show their superior performance in comparison to standard features  across  a  range  of  activity recognition  applications. 
Budiman et al.~\citep{Budiman2014Stacked} use SAEs and marginalized SAE to infer generic features in conjunction with neural networks and Extreme Learning Machines as the supervised classifiers to perform pose-based action recognition. 
Li et al.~\citep{Li2014} compare SAE, Denoising AE and PCA for unsupervised feature learning in activity recognition using smartphone sensors. They show that traditional features perform worse than the generic features inferred through autoencoders. 
Jokanovic et al.~\citep{jokanovic2016radar} use SAE to learn generic lower dimensional features and use softmax regression classifier to identify falls using radar signals. 
Other researchers~\citep{jankowski2015deep,wang2016recognition} have used AEs to reduce the dimensionality of domain specific features prior to applying traditional supervised classification models or deep belief networks.

AEs have also been extensively used in anomaly detection. Japkowicz et al.~\citep{japkowicz1995novelty} present the use of AE for novelty detection. For  noiseless data, they propose to use a reduced percentage of maximum of reconstruction error as a threshold to identify outliers. For  noisy data, they propose to identify both the intermediate positive and negative regions and subsequently optimizing the threshold until a desired accuracy is achieved. 
Manevitz and Yousef~\citep{Manevitz20071466} present an AE approach to filter documents and report better performance than traditional classifiers. They report to carry out certain type of uniform transformation before training the network to improve the performance. They discuss that choosing an appropriate threshold to identify normal documents is challenging and present several variants.
 The method that worked the best in their application is to tighten the threshold sufficiently to disallow the classification of the highest $25$ percentile error cases from the training set. 
Erfani et al.~\citep{Erfani2016121} present a hybrid approach to combine the AEs and one-class SVM ($OSVM$) for anomaly detection in high-dimensional and large-scale applications. They first extract generic features using SAE and train an $OSVM$ with linear kernel on learned features from SAE. They also use SAE as a one-class classifier by setting the threshold to be $3$ times of standard deviation away from the mean. Their results show comparable results in comparison to AE based anomaly classifier but the training and testing time greatly reduced. Sakurda and Yairi~\citep{sakurada2014anomaly} show the use of AE in anomaly detection task and compare it with PCA and Kernel PCA. They demonstrate that the AE can detect subtle anomalies that PCA could not and is less complex than Kernel PCA. 

Ensembles of AE have been used to learn diverse feature representations, mainly in the supervised settings. Ithapu et al.~\citep{ithapu2014randomized} present an ensemble of SAE by presenting it with randomized inputs and randomized sample sets of hyper-parameters from a given hyper–parameter space. They show that their approach is more accurately related to different stages of Alzheimer's disease and leads to efficient clinical trials with very less sample estimates. 
Reeve and Gavin~\citep{Reeve2015Modular} present a modular AE approach that consists of $M$ AE modules trained separately on different data representations and the combined result is defined by taking an  average of all the modules present. Their results on several benchmark datasets show improved performance in comparison to baseline of bootstrap version of the AE. Dong and Japkowicz~\citep{Dong2016Threaded} present a supervised and unsupervised ensemble approach for stream learning that uses multi-layer neural networks and AE. They train their models from multi-threads which evolve with data streams, the ensemble of the AE is trained using only the data from positive class and is accurate when anomalous training data are rare. Their method performs better as compared to the state-of-the-art in terms of detection accuracy and training time for the datasets.


The research on using AE show that it can successfully learn generic features from raw sensor data for activity and fall recognition tasks. We observe that AE can be effectively used for anomaly detection tasks and their ensembles can perform better than a single AE. In this paper, fall detection problem is formulated as an OCC or anomaly detection, where abundant data for normal activities is available during training and none for falls. We investigate the utility of features learned through AE and their ensembles for the task of fall detection. 


\section{Autoencoder Ensemble for Detecting Unseen Falls}
\label{sec:monolithic}
In the absence of training data for falls, a fall can be detected by training an AE/SAE on only the normal activities to learn generic features from a wearable device. These features can be fed to standard OCC algorithms to detect a test sequence as a normal activity or not (a fall in our case). Alternatively, based on the training data, a threshold can be set on the reconstruction error of the AE/SAE to identify a test sequence as an abnormal activity (a fall in our case) if its reconstruction error is higher than a given threshold. Intuitively, this would mean that the test sequence is very different from the training data comprising of normal activities. Below, we discuss two types of AE approaches used in the paper.

\subsection{Monolithic Autoencoders}

\begin{figure}[!ht]
\centering
\includegraphics[width=7cm,height=3cm]{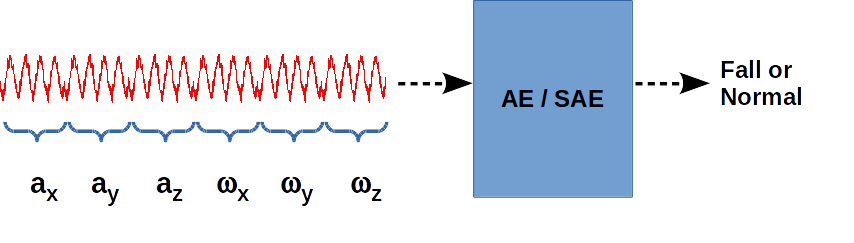}
\caption{Monolithic AE for detecting unseen falls.}
\label{fig:mono}
\end{figure}

Figure \ref{fig:mono} shows the AE/SAE for training normal activities using raw sensor data from a three-channel accelerometer and gyroscope. The raw sensor readings coming from each of the channels of accelerometer ($a_x,a_y,a_z$) and gyroscope ($\omega_x, \omega_y,\omega_z$) are combined and presented as input to the AE/SAE. For a sliding window of a fixed length ($n$ samples), $\mathbf{a_x} = [a_x^1, a_x^2, ..., a_x^n]$, $\mathbf{a_y} = [a_y^1, a_y^2, ..., a_y^n]$, and so on. The feature vector for a time window is constructed by concatenating these sensor readings as $\mathbf{f} = [\mathbf{a_x}, \mathbf{a_y}, \mathbf{a_z}, \bm{\omega_x}, \bm{\omega_y}, \bm{\omega_z}]^T$. 
We call this feature learning approach as \textit{monolithic} because it combines raw sensor data from different channels as one input to an AE.

\begin{figure}[!ht]
\centering
  \begin{subfigure}[b]{0.5\textwidth}
\centering
\includegraphics[width=8cm,height=6cm]{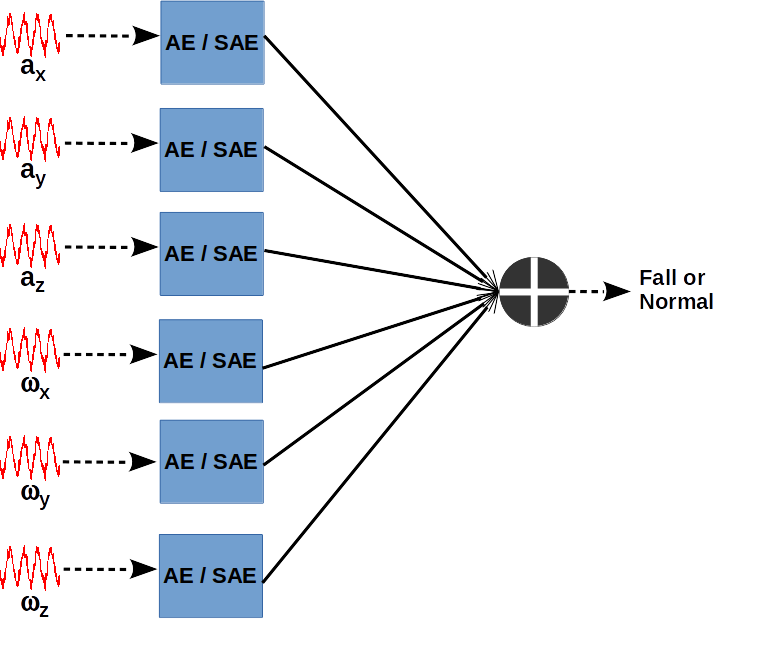}
\caption{Ensemble of $6$ separate channels of accelerometer and gyroscope.}
\label{fig:6channel}
\end{subfigure}
  \begin{subfigure}[b]{0.5\textwidth}
\centering
\includegraphics[width=8cm,height=6cm]{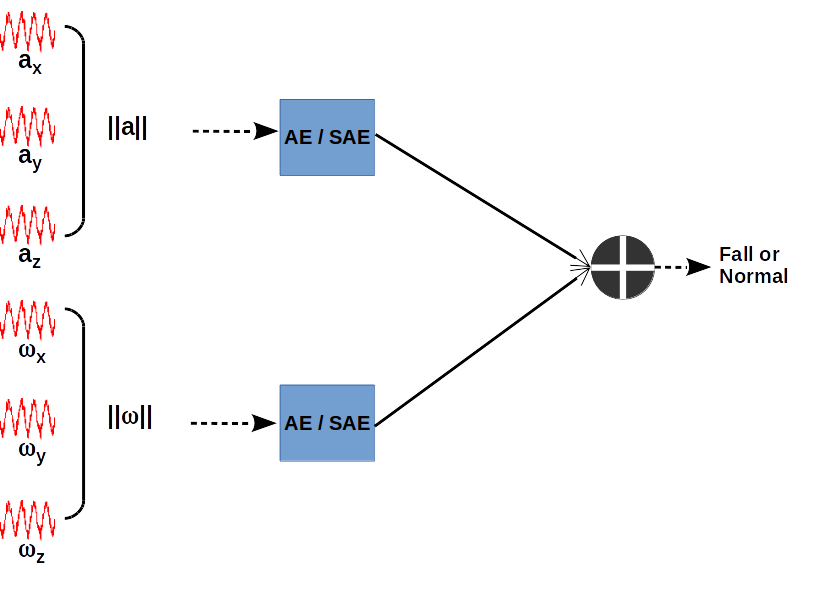}
\caption{Ensemble of $2$ channels of the magnitude of accelerometer and gyroscope.}
\label{fig:2channel}
\end{subfigure}
\caption{Channel-wise AE for detecting unseen falls.}
\label{fig:ens}
\end{figure}

\subsection{Channel-wise Autoencoders}

Li et al.~\citep{Li2014} present the use of ensemble of SAE by extracting generic features per each of the three accelerometer channels and additional channel for the magnitude of the accelerometer vector in $3$-dimensions. They extract fixed number of features for each of these $4$ channels and concatenate them. Supervised classification methods are then used on these extracted features. 
This setting can work for supervised classification but not in OCC scenario. 
In our case, we deal with OCC scenario with only normal data available during training. Therefore, separate AEs are trained on the raw data from different channels of the sensors. Each AE can detect a test sample as an unseen fall or not based on the reconstruction error and their overall result is combined to take a final decision.
We propose to use two types of channel-wise ensemble strategies for detecting unseen falls as follows:

\begin{itemize}
\item Six Channel Ensemble ($6\_CE$): 
 For each of the $6$ channels of an accelerometer and a gyroscope (i.e., $\mathbf{a_x},\mathbf{a_y},\mathbf{a_z}, \bm{\omega_x}, \bm{\omega_y},\bm{\omega_z}$), $6$ separate AE/SAE are trained 
to learn a compact representation for each channel. A decision threshold can be employed on each of these $6$ AEs to decide whether a test sample is normal or a fall. 

\item Two Channel Ensemble ($2\_CE$):
 Alternatively, we can compute the magnitude of the $3$ accelerometer channels and that of the $3$ gyroscope channels. 
The magnitude vector gives direction invariant information. 
We train two separate AE/SAE to learn a compact representation for each of the two magnitude channels. Thresholding the reconstruction error on these two channels can be used to decide whether a test sample is normal activity or a fall.

\end{itemize}

For a given test sample, the $6\_CE$ will give $6$ different decisions and the $2\_CE$ give $2$ decisions. These decisions can be combined by majority voting to arrive at a final decision; as a convention, ties are considered as falls. For simplicity, we keep the hyper-parameters for each AE/SAE corresponding to a channel as the same. 
The ensemble approach can be faster than the monolithic approach because AE/SAE per channel uses less amount of data in comparison to the combined $6$ channel data to a single AE/SAE. Figure \ref{fig:ens} shows the graphical representation of the $6\_CE$ and $2\_CE$ approaches.


\section{Optimizing the Threshold on the Reconstruction Error }
\label{sec:threshold}
For the fall detection problem, we assume that fall data is rare and is not present during training phase \cite{khan2017detecting}. Therefore, we train monlithinc and channel-wise AE and SAE on the raw sensor data to learn a compact representation of the normal activities. The next step is to identify a test sample as normal or fall based on the trained AE/SAEs.
The typical approach to identify a fall as an anomaly is to set a threshold on the reconstruction error. This threshold is generally set as the maximum of the reconstruction error on the full training data. We call this threshold \textit{MaxRE}. During testing, any sample that has a reconstruction error greater than this value can be identified as a fall. However, sensor readings are not perfect and may contain spurious data~\citep{khan2014iwaal}, which can affect this threshold. Due to the presence of a few outliers in the training data, \textit{MaxRE} is often too large, which could result in many of the falls being missed during testing time. To handle this situation, tightening of threshold is often required (as discussed in Section \ref{sec:related}). We use the approach of Erfani et al.~\citep{Erfani2016121} that sets the threshold as $3$ standard deviations away from the mean of the training data reconstruction error. We call this threshold method as \textit{StdRE}. The \textit{StdRE} threshold can result in identifying more falls during testing in comparison to \textit{MaxRE} at the cost of few false alarms because the threshold in this case is smaller in comparison to \textit{MaxRE}. A problem with \textit{StdRE} is that it is chosen in an adhoc manner and it may not be an appropriate choice for a given data.

We now present two new approaches to tighten the threshold on reconstruction error. These approaches derive the threshold from the training data such that it can better identify unseen falls. These methods are similar to finding an optimal operating point on an ROC curve by reducing false negatives at the cost of false alarms. However, in a OCC framework, it is difficult to adopt a traditional ROC approach because of the unavailability of the validation data for the negative class. The proposed methods overcome these difficulty by removing outliers from the training data prior to setting a threshold based on only the training data.

\subsection{Reduced Reconstruction Error}
\label{sec:rre}
As we discussed earlier, the raw sensor data may not be perfect and may contain spurious or incorrectly labeled readings~\citep{khan2014iwaal}. If an AE/SAE is trained on normal activities on such data, the reconstruction error for some of the samples of the training set may be very large. In this case, choosing the maximum of reconstruction error as the threshold to identify falls may lead to accepting most of the falls as normal activities. 

Khan et al.~\citep{khan2017detecting} propose to use  the concept of quartiles from descriptive statistic to remove few outliers present in the normal activities class. We use a similar idea but adapt it to AE to tighten the threshold on the reconstruction error.  We first train an AE on normal activites, then find the reconstruction error of each training sample. Given the reconstruction error on the training data comprising of only instances of normal activities, the lower quartile ($Q_1$), the upper quartile ($Q_3$) and the inter-quartile range ($IQR = Q_3 - Q_1$), a point $P$ is qualified as an outlier of the normal class, if
\begin{equation}
  P > Q_3 + \Omega \times IQR \quad || \quad P < Q_1 - \Omega \times IQR 
\label{eqn:iqr}
\end{equation}

where $\Omega$ is the rejection rate that represents the percentage of data points that are within the non-extreme limits. Based on $\Omega$, the extreme values of reconstruction error that represents spurious training data can be removed and a threshold can be chosen as the maximum of the remaining reconstruction errors. We call this method as Reduced Reconstruction Error (\textit{RRE}). The value of $\Omega$ can be found experimentally or set to remove a small fraction of the normal activities data. We describe a cross-validation technique in Section \ref{sec:cv} to find \textit{RRE} from only the normal activities.

\subsection{Inlier Reconstruction Error}

In this method, we first train an AE/SAE on full normal data and then remove the corresponding anomalous training instances based on $\Omega$ from the training set (as discussed in the previous section). After this step, we are left with training data without the outlier instances. Then, we train a new AE on this reduced data comprising of just the inlier. The idea is that the variance of reconstruction error for such inlier data will not be too high and its maximum can serve as the new threshold. We call this method as Inlier Reconstruction Error (\textit{IRE}). 

For the channel-wise ensemble approach, each AE/SAE is trained only using the raw sensor data from a specific channel of normal activities, then various thresholds, i.e., \textit{MaxRE}, \textit{StdRE}, \textit{RRE} and \textit{IRE} are computed for each channel separately. During testing, for a given threshold method, the final decision is taken as the majority voting outcome of all the AE/SAE. 
The intuition behind \textit{RRE} and \textit{IRE} is that they should provide a better trade-off between false positive and true positive rate in comparison to \textit{MaxRE} and \textit{StdRE}. The threshold \textit{MaxRE} may work better in one direction, whereas \textit{StdRE} is an adhoc approach to minimize errors. Both \textit{RRE} and \textit{IRE} are derived from the data and not arbitrarily set to a fixed number. The proposed threshold tightening methods, i.e. \textit{RRE} and \textit{IRE}, attempt to find a threshold after removing spurious sensor data from the normal activities, this may lead to improved sensitivity in detecting falls.

\subsection{Cross Validation}
\label{sec:cv}
The parameter $\Omega$ to tighten the threshold for \textit{RRE} and \textit{IRE} cannot be directly optimized because there is no validation set due to the absence of fall data during training. Khan et al. \cite{khan2017detecting} propose to remove some outliers from the normal data and consider them as proxy for unseen falls. They show that rejected outliers from the normal activities can be used to create a validation set and tune parameters of a learning algorithm in the absence of fall data. They use hidden markov models and show that some of the proxy for falls bear resemblance to actual falls. We modify this idea with respect to the autoencoders and present a cross-validation method to optimize $\Omega$ in our setting.

Firstly, we train an AE on full normal data and compute reconstruction error of each training example. Then we reject some instances from the normal activities based on a parameter $\rho$, using their reconstruction error. The parameter $\rho$ also uses IQR technique (as discussed in Section \ref{sec:rre}); however it is very different from the parameter $\Omega$. The parameter $\rho$ represents the amount of outlier data removed from normal activities to generate samples for proxy falls to create a validation set. The remaining normal activities are called non-falls. The parameter $\Omega$ represents the amount of reconstruction error removed to set a `threshold' to identify unseen falls during testing. For a given value of $\rho$, several values of $\Omega$ can be tested and the best is used for further analysis. Therefore, $\rho$ is considered as a hyper-parameter and $\Omega$ as a parameter to find \textit{RRE} and \textit{IRE}.
Then the data from both the classes (non-falls and proxy fall) is divided into $K$-folds. The non-fall data from $(K-1)$ folds is combined and an AE is trained on it. The data from $K^{th}$ fold for non-fall and proxy fall is used for testing and tuning the parameters. The process is repeated $K$ times for different values of $\Omega$, the one with the best average performance over $K$ folds is chosen for further analysis. The peformance metric is discussed in Section \ref{sec:perf}.
Lastly, for a given $\rho$, we retrain on the non-fall data. The maximum of the reconstruction error corresponding to the best $\Omega$ (obtained in the step discussed above for a given $\rho$) is taken as \textit{RRE}. To compute \textit{IRE}, we remove the outliers from the non-falls corresponding to $\Omega$; then retrain on the reduced training set and take the maximum of the reconstruction error as \textit{IRE}.

The value of hyper-parameter $\rho$ can be varied to observe an overall effect on the performance of the proposed threshold tightening methods, \textit{RRE} and \textit{IRE}. Intuitively, a large value of $\rho$ means less number of instances are removed from normal activities as proxy for fall, which may lead to classify a lot of test samples as normal activities and may miss to identify some falls. Whereas, a small value of $\rho$ means more instances from normal activities may be rejected as proxy for falls; thus, the normal class will be smaller that may result in identifying most of the falls but at the cost of more false alarms. In summary, we exect that with increase in $\rho$, both the true positive rate and false positive rate should reduce (fall is the positive class). By varying $\rho$, we can find an optimum range of operation with a good balance of true positives and false positives. It is to be noted that in this cross-validation method, no actual falls from the training set are used because it is only comprised of normal activities and all the parameters are tuned in the absence of actual falls.

\section{Experimental Analysis}
\label{sec:results}

\subsection{Performance Metrics}
\label{sec:perf}
We consider a case for detecting falls where they are not available during training and occur only during testing. Therefore, during the testing phase, we expect a skewed distribution of falls. Hence, the standard performance metrics such as accuracy may not be appropriate because it may give an over-estimated view of the performance of the classifier. To deal such a case, we use the geometric mean ($gmean$)~\citep{kubat1997addressing,khan2014iwaal} as the performance metric to present the test results and optimize the parameters during cross-validation. $gmean$ is defined as the square root of the multiplication of true positive and true negative rate, i.e.
\begin{equation}
\begin{split}
 gmean &= \sqrt{TPR * TNR} \\
 gmean &= \sqrt{TPR * (1-FPR)}
\end{split}
\end{equation}

where $TPR$ is the true positive rate, $TNR$ is the true negative rate and $FPR$ is the false positive rate. The value of $gmean$ varies from $0$ to $1$, where a $1$ means a perfect classification among falls and normal activities and $0$ as the worst outcome. We also use the $TPR$ and $FPR$ as other performance metrics to further elaborate our results.

To evaluate the performance of the proposed approaches for fall detection, we perform leave-one-subject-out cross validation (LOOCV) \cite{he2009activity}, where \emph{only} normal activities from $(N-1)$ subjects are used to train the classifiers and the $N^{th}$ subject's normal activities and fall events are used for testing. This process is repeated $N$ times and the average performance metrics are reported. This evaluation is person independent and demonstrates the generalization capabilities as the subject who is being tested is not included in training the classifiers.  

\subsection{Datasets}

We show our results on two activity recognition datasets that includes different normal activities and fall events collected via wearable devices.

\subsubsection{German Aerospace Center (DLR)~\citep{dlr65511}} This dataset is collected using an Inertial Measurement Unit with a sampling frequency of $100$ Hz. The dataset contains samples from 19 people of both genders of different age groups. The data is recorded in indoor and outdoor environments under semi-natural conditions. The sensor is placed on the belt either on the right or the left side of the body or in the right pocket in different orientations. The dataset contains labelled data of the following $7$ activities: Standing, Sitting, Lying, Walking (up/downstairs, horizontal), Running/Jogging, Jumping and Falling. 
  One of the subjects did not perform fall activity; therefore, their data is omitted from the analysis.
  
\subsubsection{Coventry Dataset (COV)~\citep{Ojetol2015data}} This dataset is collected using two SHIMMER\texttrademark sensor nodes strapped to the chest and thighs of subjects with a sampling frequency of $100$ Hz. Two protocols were followed to collect data from subjects. In Protocol 1, data for six types of fall scenarios are captured (forward, backward, right, left, real fall-backward and real fall forward) and a set of ADL (standing, lying, sitting on a chair or bed, walking, crouching, near falls and lying).  Protocol 2 involved ascending and descending stairs. $42$ young healthy individuals simulated various ADL and fall scenarios ($32$ in Protocol $1$ and $10$ in Protocol $2$). 
 These data from different types of falls are joined together to make one separate class for falls. The subjects for Protocol $2$ did not record corresponding fall data; therefore, that data is not used. In our analysis, we used accelerometer and gyroscope data from the sensor node strapped to the chest.

\subsection{Experimental Setup} 
For both the datasets, all the normal activitis are joined together to form a normal class. For COV dataset, different types of falls are joined to make a fall class.
The raw sensor data is processed using a $50\%$ overlapping sliding window. The time window size is set to $1.28$ seconds for the DLR dataset and $2.56$ seconds for the COV dataset (as shown in Khan et al.~\citep{khan2017detecting}). After pre-processing, the  DLR  dataset has  26576  normal  activities  and  84  fall  segments, and the COV dataset has 12392 normal activities and 908 fall segments. 

We test two types of AE for the analysis; one with a single hidden layer and other with three layered SAE. For the monolithic AE, the raw data within a time window for each of the $3$ channels of accelerometer and gyroscope is concatenated, which leads to $768 (=128\times6)$ input layer neurons for the DLR dataset and $1536 (=256\times6)$ input layer neurons for the COV dataset. The number of hidden neurons (i.e., the number of generic features learned) is set to $31$ (as suggested for the engineered features case in the work of Khan et al.~\citep{khan2017detecting}).
For the monolithic SAE, the number of hidden neurons in the first layer is chosen to be half of the number of input neurons, i.e. $384$ for DLR dataset and $768$ for COV dataset and the second layer has $31$ number of features. For the channel-wise ensemble method, each channels is fed to the AE/SAE separately. Therefore, the number of neurons in the input layer per AE is set to $128$ for DLR dataset and $256$ for COV dataset and the hidden layers has $31$ neurons. For the channel-wise SAE, the hidden neurons for first layer is half the number of input layer, i.e. $64$ for DLR dataset and $128$ for the COV dataset. The second hidden layer for both the datasets has $31$ neurons. The number of training epochs is fixed to $10$ for all the different autoencdoers. Rest of the parameters such as the sparsity parameter, activation function etc., are kept at the default values~\citep{website:ae}. Compressed features learned through monolithic AE and SAE are further used to train $OSVM$ and One-class nearest neighbour ($OCNN$) classifiers~\citep{Khan:KER:2014} for comparison. 

\subsubsection{Internal Cross-Validation}

For $OCNN$, the number of nearest neighbours to identify an outlier is kept as $1$. $OSVM$ has a parameter $\nu$ (or the outlier fraction), which is the expected proportion of outliers in the training data. The value of this parameter is tuned, similar to parameter optimization discussed in Section \ref{sec:cv}. That is, reject a small portion of normal class data as a proxy for unseen falls for a given $\rho$ and create a validation set. Then perform a $K$ fold cross-validation for different values of $\nu$ and choose the one with the largest average $gmean$ over all the $K$-folds. The `KernelScale' parameter is set to `auto' and 'Standardize' to `true', other parameters are kept to default values \cite{website:svm}.

An internal $K=3$-fold cross validation is employed to optimize the parameters $\nu$ for the $OSVM$ and $\Omega$ for the \textit{RRE} and \textit{IRE} thresholding methods. The parameter $\Omega$ is varied from from $[0.001, 0.01, 0.1, 0.5, 1, 1.5, 1.7239, 2, 2.5, 3]$ and $\nu$ is varied from $[0.1,0.3, 0.5, 0.7, 0.9]$. The best parameter is chosen based on the average $gmean$ over $K$ folds. 
To understand the effect of removing outlier data from the normal activities in building classification models for unseen falls
, the hyper-parameter $\rho$ is varied from $[0.001, 0.01, 0.1, 0.5, 1, 1.5, 1.7239, 2, 2.5, 3]$.

Along with $6$ channel raw data to train different classifiers to detect unseen falls, we also use $2$ channel magnitude data from each of the datasets to train different classifiers. Therefore, in the experiment we compare the following different classifiers for two types of channels data (i.e. $6$ and $2$ channels):
\begin{itemize}[leftmargin=*]
 \item Two types of AE, i.e. single layer AE and three layered SAE.
 \item Four types of thresholding methods i.e. \textit{MaxRE}, \textit{StdRE}, \textit{RRE} and \textit{IRE}.
 \item Two types of feature learning techniques - (i) monolithic and (ii) channel-wise ensemble.
 \item Two one-class classifiers ($OCNN$ and $OSVM$ ) trained on features learned from AE and SAE (not for the channel-wise case).
\end{itemize}

This results in $20$ different classifiers trained per $6$ / $2$ channels input raw sensor data, we compare their performance in the next section.

\subsection{Results and Discussion}

Tables \ref{tab:DLR6} and \ref{tab:DLR2} show the results for the DLR datasets for $6$ and $2$ channel input raw data. The results correspond to $\rho=1.5$. Tables \ref{tab:DLR6occ} and \ref{tab:DLR2occ} show the results when the features learned using AE and SAE are fed to OSVM and OCNN. We observe that for the $6$ channel case, the best $gmean$ is obtained for channel-wise AE with \textit{RRE} followed by \textit{IRE} method (see Tables \ref{tab:DLR6gmean}. The traditional thresholding methods of \textit{MaxRE} and \textit{StdRE} does not perform well. We observe similar results for DLR dataset with $2$ channel input; however, the $gmean$ values using $6$ channel input data are higher. 
Both the \textit{RRE} and \textit{IRE} methods with channel-wise AE give good trade-off between $TPR$ (see Tables \ref{tab:DLR6TPR} and \ref{tab:DLR2TPR}) and $FPR$ (see Tables \ref{tab:DLR6FPR} and \ref{tab:DLR2FPR}). 

Results on the COV dataset for the $6$ and $2$ channels input data are shown in Table \ref{tab:COV6} and \ref{tab:COV2}. For the $6$ channel case, channel-wise ensemble method with AE for \textit{RRE} and \textit{IRE} give equivalent values of $gmean$, which is higher than other methods of thresholding. Both the best methods give a good trade-off between $TPR$ and $FPR$ (see Tables \ref{tab:COV6TPR} and \ref{tab:COV6FPR}).
 For the $2$ channel case, the \textit{IRE} threshold method for both the monolithic and channel-wise approaches for AE and SAE give equivalent performance along with monolithic SAE with \textit{RRE}. The channel-wise approach gives more false alarms but detects more falls than the monolithic approach.
For both the DLR and COV datasets, the OCNN classifier perform worse than the proposed methods  because it gives large number of false alarms; whereas, OSVM classifies all the test samples as falls (see Tables \ref{tab:DLR6occ}, \ref{tab:DLR2occ}, \ref{tab:COV6occ}, \ref{tab:COV2occ}).

By convention, we classify a test sample as a fall in case of a tie in the channel-wise approach. The probability of a tie occurring is higher in $2$ channel ensemble method than in $6$ channel ensemble; therefore, its sensitivity to detect falls is higher than the $6$ channel case with an increase in the false alarm rate. We observe this behavior for both the DLR and COV datasets (see the Channel-wise rows in Tables \ref{tab:DLR6FPR}, \ref{tab:DLR2FPR}, and \ref{tab:COV6FPR}, \ref{tab:COV2FPR} ).
From this experiment we infer that the traditional methods of thresholding, i.e., \textit{MaxRE} and \textit{StdRE}, are not suitable for the task of fall detection. \textit{MaxRE} may not work properly because of the presence of noise in the sensor data that can significantly increase the reconstruction error of an AE/SAE, leading to classify most of the test samples as normal activity. The \textit{StdRE} is an ad-hoc approach that arbitrarily chooses a threshold to identify fall and does not derive it from a given dataset. However, it can perform better than \textit{MaxRE}, in terms of identifying more falls. 
Both of these methods attempt to find a discriminating threshold from the training dataset to to get a good trade-off between $TPR$ and $FPR$.
Our experiments suggest that for both the datasets, the proposed threshold tightening methods \textit{RRE} and \textit{IRE} with channel-wise ensemble approach perform equivalently and consistently better than the traditional methods of threshold tightening.

\begin{table}[!ht]
    \begin{subtable}{.8\linewidth}
  \centering
\caption{$gmean$ values.}
    \begin{tabular}{|p{1.5cm}|p{2.1cm}|l|l|l|l|}\hline
\textbf{Features Types} & \textbf{Autoencoder Type} &\multicolumn{4}{c|}{\textbf{Thresholding}}\\ \cline{3-6}
                            &   & \textbf{$\textit{MaxRE}$} & \textbf{$\textit{StdRE}$}   & \textbf{$\textit{RRE}$} & \textbf{$\textit{IRE}$} \\ \hline\hline
                            \multirow{2}{*}{Monolithic}    & AE  & 0  & 0.106 & 0.825 & 0.757 \\ \cline{2-6}
                                                           & SAE & 0  & 0.234 & 0.840 & 0.837 \\ \hline \hline
                            Channel-  & AE  & 0	& 0.547  & \cellcolor{Gray}{0.860} & \cellcolor{Gray}{0.849} \\ \cline{2-6}
                            wise      & SAE & 0	& 0.334  & 0.818 & 0.811 \\ \hline 
    \end{tabular}
    \label{tab:DLR6gmean}
    \end{subtable}
    \\
    \begin{subtable}{.8\linewidth}  
    \vspace{5mm}
\caption{$TPR$ values.}
    \begin{tabular}{|p{1.5cm}|p{2.1cm}|l|l|l|l|}\hline
\textbf{Features Types} & \textbf{Autoencoder Type} &\multicolumn{4}{c|}{\textbf{Thresholding}}\\ \cline{3-6}
                            &   & \textbf{$\textit{MaxRE}$} & \textbf{$\textit{StdRE}$}   & \textbf{$\textit{RRE}$} & \textbf{$\textit{IRE}$} \\ \hline\hline
                            \multirow{2}{*}{Monolithic}    & AE  & 0  & 0.056 & 0.856 & 0.762 \\ \cline{2-6}
                                                           & SAE & 0  & 0.138 & 0.893 & 0.893 \\ \hline \hline
                            Channel-  & AE  & 0  & 0.428 &  \cellcolor{Gray}{0.902} & \cellcolor{Gray}{0.840} \\ \cline{2-6}
                            wise      & SAE & 0  & 0.226 & 0.774 & 0.750 \\ \hline 
    \end{tabular}
    \label{tab:DLR6TPR}
    \end{subtable}
    \\
    \begin{subtable}{.8\linewidth}    
    \vspace{5mm}
    \caption{$FPR$ values.}
\begin{tabular}{|p{1.5cm}|p{2.1cm}|l|l|l|l|}\hline
\textbf{Features Types} & \textbf{Autoencoder Type} &\multicolumn{4}{c|}{\textbf{Thresholding}}\\ \cline{3-6}
                            &   & \textbf{$\textit{MaxRE}$} & \textbf{$\textit{StdRE}$}   & \textbf{$\textit{RRE}$} & \textbf{$\textit{IRE}$} \\ \hline\hline
                            \multirow{2}{*}{Monolithic}    & AE  & 5.9e-6  & 0.025 & 0.189 & 0.169 \\ \cline{2-6}
                                                           & SAE & 5.9e-6  & 0.025 & 0.199 & 0.204 \\ \hline \hline
                            Channel-  & AE  & 0  & 0.010 & \cellcolor{Gray}{0.169} & \cellcolor{Gray}{0.122} \\ \cline{2-6}
                            wise      & SAE & 0  & 0.008 &  0.088 & 0.079 \\ \hline 
    \end{tabular}
    \label{tab:DLR6FPR}
    \end{subtable}
    \\
    \begin{subtable}{.8\linewidth}    
    \vspace{5mm}
\caption{Performance on OSVM and OCNN methods.}
\begin{tabular}{|p{1.5cm}|p{2.1cm}|l|l|l|l|}\hline
\textbf{Classifier} & \textbf{Autoencoder Type} &$gmean$ & $TPR$ & $FPR$\\ \hline
\multirow{2}{*}{OSVM} & AE & 0 & 1 & 1 \\ \cline{2-5}
		      &SAE & 0 & 1 & 1 \\ \hline \hline
\multirow{2}{*}{OCNN} & AE & 0.460 & 0.911 & 0.761 \\ \cline{2-5}
		      &SAE & 0.423 & 0.681 & 0.701 \\ \hline
\end{tabular}
\label{tab:DLR6occ}
\end{subtable}		      
\caption{Performance of different fall detection methods on DLR dataset ($6$ channels) for $\rho=1.5$}
\label{tab:DLR6}
\end{table}

\begin{table}[!ht]
    \begin{subtable}{.8\linewidth}
  \centering
    \caption{$gmean$ values.}
\begin{tabular}{|p{1.5cm}|p{2.1cm}|l|l|l|l|}\hline
\textbf{Features Types} & \textbf{Autoencoder Type} &\multicolumn{4}{c|}{\textbf{Thresholding}}\\ \cline{3-6}
                            &   & \textbf{$\textit{MaxRE}$} & \textbf{$\textit{StdRE}$}   & \textbf{$\textit{RRE}$} & \textbf{$\textit{IRE}$} \\ \hline\hline
                            \multirow{2}{*}{Monolithic}    & AE  & 0  & 0 & 0.504 & 0.774 \\ \cline{2-6}
                                                           & SAE & 0  & 0 & 0.630 & 0.776 \\ \hline \hline
                            Channel-  & AE  & 0.013 & 0.487 & \cellcolor{Gray}{0.839} & \cellcolor{Gray}{0.822} \\ \cline{2-6}
                            wise      & SAE & 0	    & 0.446 & 0.678 & 0.655 \\ \hline 
    \end{tabular}
    \label{tab:DLR2gmean}
    \end{subtable}
    \\
    \begin{subtable}{.8\linewidth}    
    \vspace{5mm}
    \caption{$TPR$ values.}
\begin{tabular}{|p{1.5cm}|p{2.1cm}|l|l|l|l|}\hline
\textbf{Features Types} & \textbf{Autoencoder Type} &\multicolumn{4}{c|}{\textbf{Thresholding}}\\ \cline{3-6}
                            &   & \textbf{$\textit{MaxRE}$} & \textbf{$\textit{StdRE}$}   & \textbf{$\textit{RRE}$} & \textbf{$\textit{IRE}$} \\ \hline\hline
                            \multirow{2}{*}{Monolithic}    & AE  & 0  & 0 & 0.966 & 0.949 \\ \cline{2-6}
                                                           & SAE & 0  & 0 & 0.959 & 0.941 \\ \hline \hline
                            Channel-  & AE  & 0.003 & 0.323 & \cellcolor{Gray}{0.941} & \cellcolor{Gray}{0.926} \\ \cline{2-6}
                            wise      & SAE & 0	    & 0.329 & 0.629 & 0.579 \\ \hline 
    \end{tabular}
    \label{tab:DLR2TPR}
    \end{subtable}
    \\
    \begin{subtable}{.8\linewidth}    
    \vspace{5mm}
    \caption{$FPR$ values.}
\begin{tabular}{|p{1.5cm}|p{2.1cm}|l|l|l|l|}\hline
\textbf{Features Types} & \textbf{Autoencoder Type} &\multicolumn{4}{c|}{\textbf{Thresholding}}\\ \cline{3-6}
                            &   & \textbf{$\textit{MaxRE}$} & \textbf{$\textit{StdRE}$}   & \textbf{$\textit{RRE}$} & \textbf{$\textit{IRE}$} \\ \hline\hline
                            \multirow{2}{*}{Monolithic}    & AE  & 7.4e-5  & 0.037 & 0.705 & 0.363 \\ \cline{2-6}
                                                           & SAE & 3.7e-5  & 0.039 & 0.544 & 0.353 \\ \hline \hline
                            Channel-  & AE  & 1.0e-4  & 0.032 & \cellcolor{Gray}{0.245} & \cellcolor{Gray}{0.264} \\ \cline{2-6}
                            wise      & SAE & 7.7e-5  & 0.033 &  0.099 & 0.094 \\ \hline 
    \end{tabular}
    \label{tab:DLR2FPR}
    \end{subtable}
    \\
    \begin{subtable}{.8\linewidth}    
    \vspace{5mm}
\caption{Performance on OSVM and OCNN methods.}
\begin{tabular}{|p{1.5cm}|p{2.1cm}|l|l|l|l|}\hline
\textbf{Classifier} & \textbf{Autoencoder Type} &$gmean$ & $TPR$ & $FPR$\\ \hline
\multirow{2}{*}{OSVM} & AE & 0 & 1 & 1 \\ \cline{2-5}
		      &SAE & 0 & 1 & 1 \\ \hline \hline
\multirow{2}{*}{OCNN} & AE & 0.459 & 0.815 & 0.719 \\ \cline{2-5}
		      &SAE & 0.318 & 0.317 & 0.449 \\ \hline
\end{tabular}
\label{tab:DLR2occ}
\end{subtable}		      
\caption{Performance of different fall detection methods on DLR dataset ($2$ channels) for $\rho=1.5$}
\label{tab:DLR2}
\end{table}

\begin{table}[!ht]
    \begin{subtable}{.8\linewidth}
  \centering
    \caption{$gmean$ values.}
\begin{tabular}{|p{1.5cm}|p{2.1cm}|l|l|l|l|}\hline
\textbf{Features Types} & \textbf{Autoencoder Type} &\multicolumn{4}{c|}{\textbf{Thresholding}}\\ \cline{3-6}
                            &   & \textbf{$\textit{MaxRE}$} & \textbf{$\textit{StdRE}$}   & \textbf{$\textit{RRE}$} & \textbf{$\textit{IRE}$} \\ \hline\hline
                            \multirow{2}{*}{Monolithic}    & AE  & 0.015  & 0.744 & 0.774 & 0.771 \\ \cline{2-6}
                                                           & SAE & 0.019  & 0.743 & 0.772 & 0.771 \\ \hline \hline
                            Channel-  & AE  & 0.014  & 0.463 & \cellcolor{Gray}{0.795} & \cellcolor{Gray}{0.795} \\ \cline{2-6}
                            wise      & SAE & 0      & 0.226 & 0.737 & 0.707 \\ \hline 
    \end{tabular}
    \label{tab:COV6gmean}
    \end{subtable}
    \\
    \begin{subtable}{.8\linewidth}    
    \vspace{5mm}
    \caption{$TPR$ values.}
\begin{tabular}{|p{1.5cm}|p{2.1cm}|l|l|l|l|}\hline
\textbf{Features Types} & \textbf{Autoencoder Type} &\multicolumn{4}{c|}{\textbf{Thresholding}}\\ \cline{3-6}
                            &   & \textbf{$\textit{MaxRE}$} & \textbf{$\textit{StdRE}$}   & \textbf{$\textit{RRE}$} & \textbf{$\textit{IRE}$} \\ \hline\hline
                            \multirow{2}{*}{Monolithic}    & AE  & 0.004  & 0.589 & 0.744 & 0.740 \\ \cline{2-6}
                                                           & SAE & 0.007  & 0.588 & 0.738 & 0.738 \\ \hline \hline
                            Channel-  & AE  & 0.003  & 0.248 & \cellcolor{Gray}{0.7} & \cellcolor{Gray}{0.7} \\ \cline{2-6}
                            wise      & SAE & 0      & 0.082 & 0.665 & 0.573 \\ \hline 
    \end{tabular}
    \label{tab:COV6TPR}
    \end{subtable}
    \\
    \begin{subtable}{.8\linewidth}    
    \vspace{5mm}
    \caption{$FPR$ values.}
\begin{tabular}{|p{1.5cm}|p{2.1cm}|l|l|l|l|}\hline
\textbf{Features Types} & \textbf{Autoencoder Type} &\multicolumn{4}{c|}{\textbf{Thresholding}}\\ \cline{3-6}
                            &   & \textbf{$\textit{MaxRE}$} & \textbf{$\textit{StdRE}$}   & \textbf{$\textit{RRE}$} & \textbf{$\textit{IRE}$} \\ \hline\hline
                            \multirow{2}{*}{Monolithic}    & AE  & 1.1e-4  & 0.017 & 0.169 & 0.169 \\ \cline{2-6}
                                                           & SAE & 1.1e-4  & 0.017 & 0.166 & 0.167 \\ \hline \hline
                            Channel-  & AE  & 0  & 0.002  & \cellcolor{Gray}{0.067} & \cellcolor{Gray}{0.072} \\ \cline{2-6}
                            wise      & SAE & 0  & 5.9e-5 & 0.128 & 0.078 \\ \hline 
    \end{tabular}
    \label{tab:COV6FPR}
    \end{subtable}
    \\
    \begin{subtable}{.8\linewidth}    
    \vspace{5mm}
\caption{Performance on OSVM and OCNN methods.}
\begin{tabular}{|p{1.5cm}|p{2.1cm}|l|l|l|l|}\hline
\textbf{Classifier} & \textbf{Autoencoder Type} &$gmean$ & $TPR$ & $FPR$\\ \hline
\multirow{2}{*}{OSVM} & AE & 0 & 1 & 1 \\ \cline{2-5}
		      &SAE & 0 & 1 & 1 \\ \hline \hline
\multirow{2}{*}{OCNN} & AE & 0.432 & 0.977 & 0.805 \\ \cline{2-5}
		      &SAE & 0.484 & 0.949 & 0.751 \\ \hline
\end{tabular}
\label{tab:COV6occ}
\end{subtable}		      
\caption{Performance of different fall detection methods on COV dataset ($6$ channels) for $\rho=1.5$}
\label{tab:COV6}
\end{table}

\begin{table}[!ht]
    \begin{subtable}{.8\linewidth}
  \centering
    \caption{$gmean$ values.}
\begin{tabular}{|p{1.5cm}|p{2.1cm}|l|l|l|l|}\hline
\textbf{Features Types} & \textbf{Autoencoder Type} &\multicolumn{4}{c|}{\textbf{Thresholding}}\\ \cline{3-6}
                            &   & \textbf{$\textit{MaxRE}$} & \textbf{$\textit{StdRE}$}   & \textbf{$\textit{RRE}$} & \textbf{$\textit{IRE}$} \\ \hline\hline
                            \multirow{2}{*}{Monolithic}    & AE  & 0.041  & 0.743 & 0.668 & 0.785 \\ \cline{2-6}
                                                           & SAE & 0.019  & 0.724 & 0.784 & 0.784 \\ \hline \hline
                            Channel-  & AE  & 0.337  & 0.767 & 0.726 & \cellcolor{Gray}{0.788} \\ \cline{2-6}
                            wise      & SAE & 0.331  & 0.757 & 0.739 & 0.786 \\ \hline 
    \end{tabular}
    \label{tab:COV2gmean}
    \end{subtable}
    \\
    \begin{subtable}{.8\linewidth}    
    \vspace{5mm}
    \caption{$TPR$ values.}
\begin{tabular}{|p{1.5cm}|p{2.1cm}|l|l|l|l|}\hline
\textbf{Features Types} & \textbf{Autoencoder Type} &\multicolumn{4}{c|}{\textbf{Thresholding}}\\ \cline{3-6}
                            &   & \textbf{$\textit{MaxRE}$} & \textbf{$\textit{StdRE}$}   & \textbf{$\textit{RRE}$} & \textbf{$\textit{IRE}$} \\ \hline\hline
                            \multirow{2}{*}{Monolithic}    & AE  & 0.012  & 0.587 & 0.729 & 0.698 \\ \cline{2-6}
                                                           & SAE & 0.007  & 0.557 & 0.677 & 0.674 \\ \hline \hline
                            Channel-  & AE  & 0.147  & 0.621 & 0.805 & \cellcolor{Gray}{0.779} \\ \cline{2-6}
                            wise      & SAE & 0.142  & 0.606 & 0.781 & 0.779 \\ \hline 
    \end{tabular}
    \label{tab:COV2TPR}
    \end{subtable}
    \\
    \begin{subtable}{.8\linewidth}    
    \vspace{5mm}
    \caption{$FPR$ values.}
\begin{tabular}{|p{1.5cm}|p{2.1cm}|l|l|l|l|}\hline
\textbf{Features Types} & \textbf{Autoencoder Type} &\multicolumn{4}{c|}{\textbf{Thresholding}}\\ \cline{3-6}
                            &   & \textbf{$\textit{MaxRE}$} & \textbf{$\textit{StdRE}$}   & \textbf{$\textit{RRE}$} & \textbf{$\textit{IRE}$} \\ \hline\hline
                            \multirow{2}{*}{Monolithic}    & AE  & 1.7e-4  & 0.013 & 0.287 & 0.094 \\ \cline{2-6}
                                                           & SAE & 1.7e-4  & 0.012 & 0.06 & 0.056 \\ \hline \hline
                            Channel-  & AE  & 1.1e-4  & 0.015 & 0.298 & \cellcolor{Gray}{0.186} \\ \cline{2-6}
                            wise      & SAE & 1.1e-4  & 0.016 & 0.255 & 0.182 \\ \hline 
    \end{tabular}
    \label{tab:COV2FPR}
    \end{subtable}
    \\
    \begin{subtable}{.8\linewidth}    
    \vspace{5mm}
\caption{Performance on OSVM and OCNN methods.}
\begin{tabular}{|p{1.5cm}|p{2.1cm}|l|l|l|l|}\hline
\textbf{Classifier} & \textbf{Autoencoder Type} &$gmean$ & $TPR$ & $FPR$\\ \hline
\multirow{2}{*}{OSVM} & AE & 0 & 1 & 1 \\ \cline{2-5}
		      &SAE & 0 & 1 & 1 \\ \hline \hline
\multirow{2}{*}{OCNN} & AE & 0.594 & 0.905 & 0.606 \\ \cline{2-5}
		      &SAE & 0.580 & 0.606 & 0.432 \\ \hline
\end{tabular}
\label{tab:COV2occ}
\end{subtable}		      
\caption{Performance of different fall detection methods on COV dataset ($2$ channels) for $\rho=1.5$}
\label{tab:COV2}
\end{table}


\begin{figure}[!htb]
\vspace{-23mm}
\centering
    \begin{subfigure}{0.7\textwidth}
  \centering
  \includegraphics[width=\textwidth]{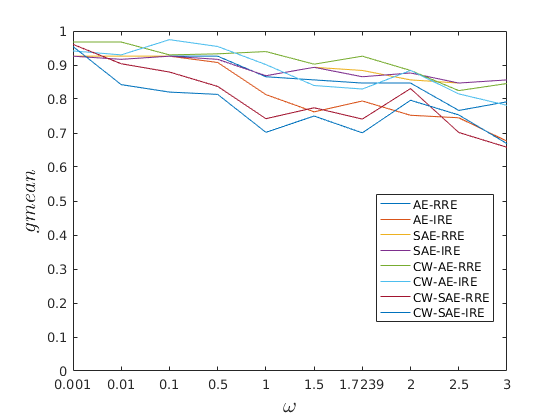} 
  \caption{True Positive Rate} 
  \label{fig:DLRTPR}
  \end{subfigure}

  \begin{subfigure}{0.7\textwidth}
  \centering
  \includegraphics[width=\textwidth]{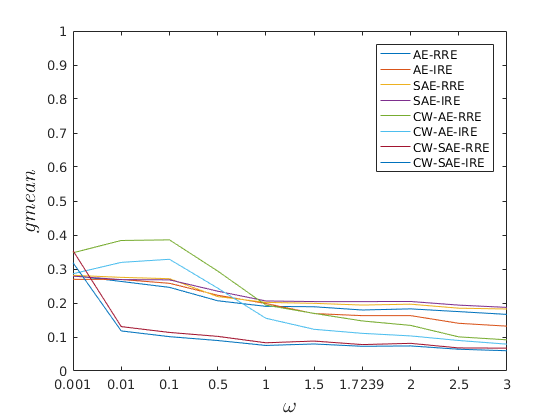} 
  \caption{False Positive Rate}
  \label{fig:DLRFPR} 
  \end{subfigure}
  \begin{subfigure}{0.7\textwidth}
  \centering
  \includegraphics[width=\textwidth]{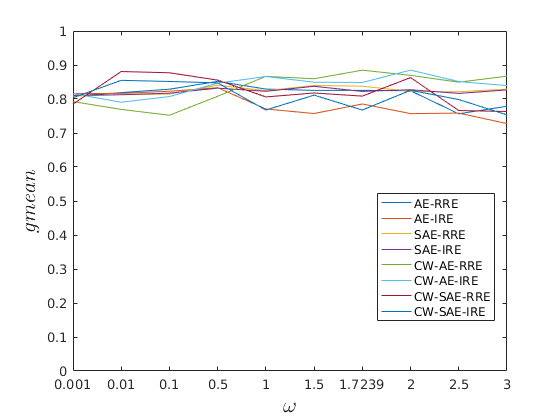} 
  \caption{$gmean$} 
  \label{fig:DLRgmean}
  \end{subfigure}
  \caption{Performance of top $5$ fall detection methods by varying $\rho$ on DLR dataset. AE - Single Layer Autoencoder, SAE - 3 Layer Stacked Autoencoder, RRE - Reduced Reconstruction Error, IRE - Inlier Reconstruction Error, CW - Channel-wise Ensemble}
  \label{fig:DLR}
\end{figure}

\begin{figure}[!htb]
\vspace{-23mm}
\centering
    \begin{subfigure}{0.7\textwidth}
  \centering
  \includegraphics[width=\textwidth]{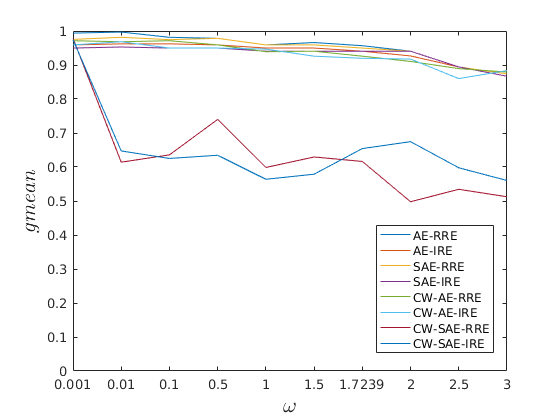} 
  \caption{True Positive Rate} 
  \label{fig:DLRTPR-norm}
  \end{subfigure}

  \begin{subfigure}{0.7\textwidth}
  \centering
  \includegraphics[width=\textwidth]{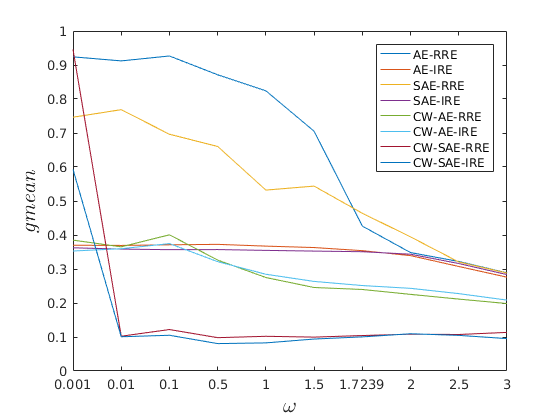} 
  \caption{False Positive Rate}
  \label{fig:DLRFPR-norm} 
  \end{subfigure}
  \begin{subfigure}{0.7\textwidth}
  \centering
  \includegraphics[width=\textwidth]{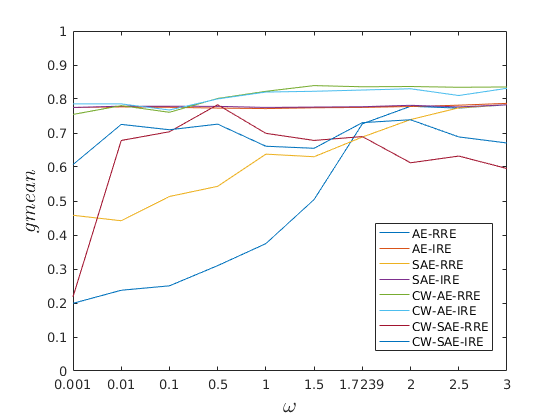} 
  \caption{$gmean$} 
  \label{fig:DLRgmean-norm}
  \end{subfigure}
  \caption{Performance of top $5$ fall detection methods by varying $\rho$ on DLR-norm dataset. AE - Single Layer Autoencoder, SAE - 3 Layer Stacked Autoencoder, RRE - Reduced Reconstruction Error, IRE - Inlier Reconstruction Error, CW - Channel-wise Ensemble}
  \label{fig:DLR-norm}
\end{figure}

\begin{figure}[htb]
\vspace{-23mm}
\centering
    \begin{subfigure}{0.7\textwidth}
  \centering
  \includegraphics[width=\textwidth]{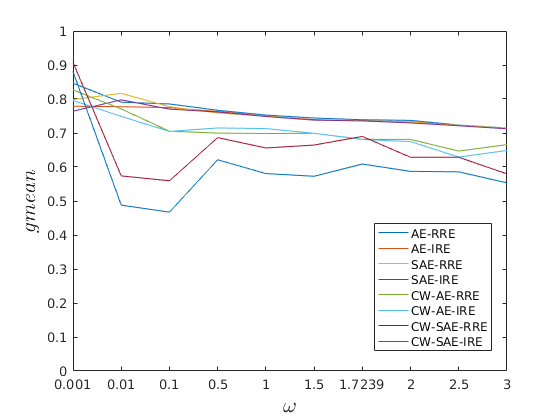} 
  \caption{True Positive Rate} 
  \label{fig:COVTPR}
  \end{subfigure}

  \begin{subfigure}{0.7\textwidth}
  \centering
  \includegraphics[width=\textwidth]{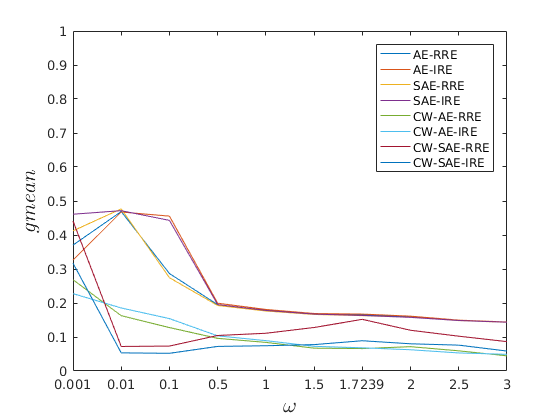} 
  \caption{False Positive Rate}
  \label{fig:COVFPR} 
  \end{subfigure}
  \begin{subfigure}{0.7\textwidth}
  \centering
  \includegraphics[width=\textwidth]{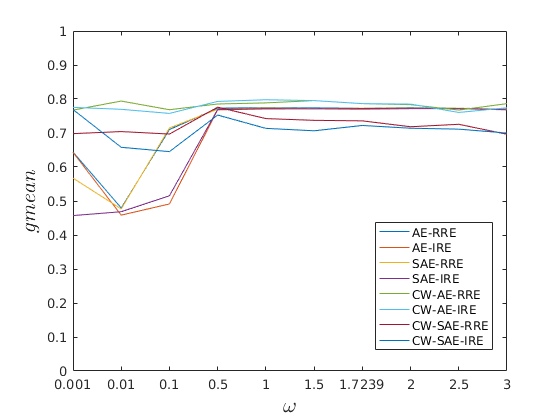} 
  \caption{$gmean$} 
  \label{fig:COVgmean}
  \end{subfigure}
  \caption{Performance of top $5$ fall detection methods by varying $\rho$ on COV dataset. AE - Single Layer Autoencoder, SAE - 3 Layer Stacked Autoencoder, RRE - Reduced Reconstruction Error, IRE - Inlier Reconstruction Error, CW - Channel-wise Ensemble}
  \label{fig:COV}
\end{figure}

\begin{figure}[htb]
\vspace{-23mm}
\centering
    \begin{subfigure}{0.7\textwidth}
  \centering
  \includegraphics[width=\textwidth]{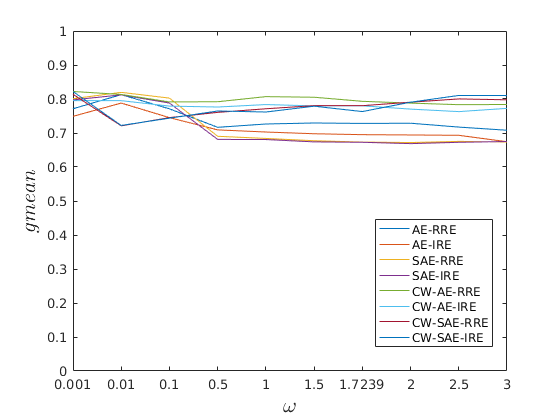} 
  \caption{True Positive Rate} 
  \label{fig:COV-normTPR}
  \end{subfigure}

  \begin{subfigure}{0.7\textwidth}
  \centering
  \includegraphics[width=\textwidth]{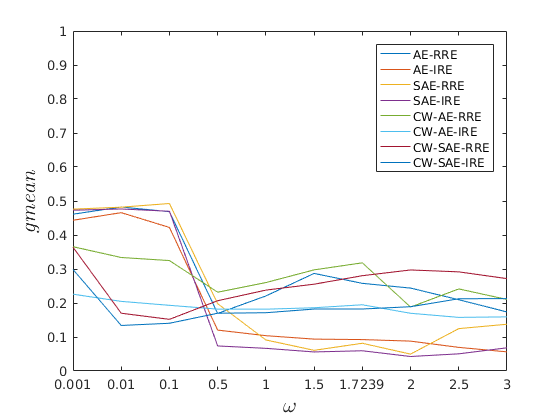} 
  \caption{False Positive Rate}
  \label{fig:COV-normFPR} 
  \end{subfigure}
  \begin{subfigure}{0.7\textwidth}
  \centering
  \includegraphics[width=\textwidth]{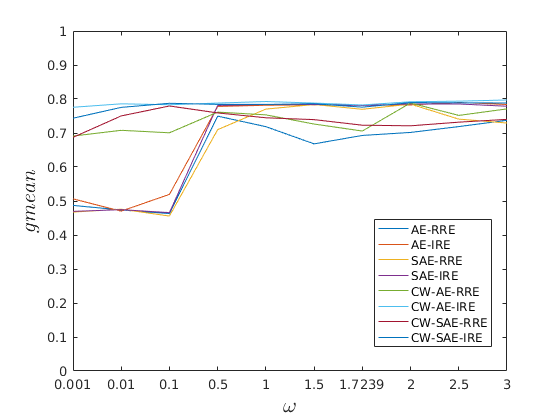} 
  \caption{$gmean$} 
  \label{fig:COV-normgmean}
  \end{subfigure}
  \caption{Performance of top $5$ fall detection methods by varying $\rho$ on COV-norm dataset. AE - Single Layer Autoencoder, SAE - 3 Layer Stacked Autoencoder, RRE - Reduced Reconstruction Error, IRE - Inlier Reconstruction Error, CW - Channel-wise Ensemble}
  \label{fig:COV-norm}
\end{figure}

We vary the hyper-parameter $\rho$ to understand its impact on the performance of different thresholding techniques. Figures \ref{fig:DLR} and \ref{fig:DLR-norm} show the variation of $TPR$, $FPR$ and $gmean$ with increasing $\rho$. We observe that as the value of $\rho$ increases, both $TPR$ and $FPR$ reduce. The reason is that at smaller values of $\rho$, a large portion of normal data is rejected as outliers and used for parameter tuning; thus, the number of instances in the non-fall class is small. This means that the AE/SAE will learn on a smaller dataset and will reject most of the variations from this small subset of normal activities as potential falls. Consequently, many falls will also be identified correctly. The reverse behavior will happen when $\rho$ is large, thus less number of normal data is rejected as outliers and the class of normal activities will be large. 
This will reduce the number of false alarms but can also lead to missing to identify some falls. The experimental observation for each of the $6$ and $2$ channel datasets is consistent  with this intuition discussed in Section \ref{sec:cv}. Similar observation can be made for the COV dataset from Figures \ref{fig:COV} and \ref{fig:COV-norm}. For both datasets, we notice that at large value of $\rho$, the performance of best thresholding approaches drops slower.
This experimental observation suggests that a smaller amount of data (corresponding to $\rho \ge 1.5$) may be removed from the normal activities class as outliers, which can be used as a validation set to optimize the parameters of the AE/SAE and better performance can be achieved for identifying unseen falls. We also infer that channel-wise approach outperforms monolithic in all the $6$ and $2$ channel data variants of both the datasets. 

\section{Conclusions and Future Work}
\label{sec:conclusions}
A fall is a rare event; therefore, it is difficult to build classification models using traditional supervised algorithms in the absence of training data.
An associated challenge for fall detection problem is to extract discriminative features in the absence of fall data for training generalizable classifiers. In this paper, we presented solutions to deal with these issues. Firstly, we formulated a fall detection problem as a one-class classification or outlier detection problem. Secondly, we presented the use of AE, more specifically a novel way to train separate AE for each channel of the wearable sensor, to learn generic features and create their ensemble. We proposed threshold tightening methods to identify unseen falls accurately. 
This work provides useful insights that an ensemble based on channels of a wearable device with optimized threshold is a useful technique to identify unseen falls.
In future work, we are exploring extreme value theory and combining it with the proposed approaches to identify unseen falls.

\section*{Acknowledgments}
This work was partially supported by the AGE-WELL NCE Trainee Award Program and by the Canadian Consortium on Neurodegeneration in Aging (CCNA). 

\section*{References}
\bibliographystyle{elsarticle-harv}
\bibliography{references}

\end{document}